\documentclass[conference]{IEEEtran}
\usepackage{times}

\usepackage[numbers]{natbib}
\usepackage{multicol}
\usepackage[bookmarks=true]{hyperref}

\usepackage{amssymb}
\usepackage{amsmath}
\usepackage{amsfonts}
\usepackage{bm}
\usepackage{bbm}
\usepackage{subcaption}
\usepackage{multirow}
\usepackage{graphicx}
\usepackage{booktabs}

\def\eqref#1{equation~\ref{#1}}

\def\1{\bm{1}}

\def\eps{{\epsilon}}

\def\vone{{\bm{1}}}

\def\vtau{{\bm{\tau}}}

\def\vq{{\bm{q}}}

\def\vv{{\bm{v}}}

\DeclareMathAlphabet{\mathsfit}{\encodingdefault}{\sfdefault}{m}{sl}
\SetMathAlphabet{\mathsfit}{bold}{\encodingdefault}{\sfdefault}{bx}{n}

\newcommand{\E}{\mathbb{E}}

\newcommand{\R}{\mathbb{R}}

\DeclareMathOperator*{\argmax}{arg\,max}

\begin{document}

\title{Quantile QT-Opt for Risk-Aware Vision-Based Robotic Grasping}




%
\author{\authorblockN{Cristian Bodnar\authorrefmark{1},
Adrian Li\authorrefmark{2},
Karol Hausman\authorrefmark{3}, 
Peter Pastor\authorrefmark{2} and
Mrinal Kalakrishnan\authorrefmark{2}}
\authorblockA{\authorrefmark{1}Department of Computer Science and Technology \\ University of Cambridge, Cambridge, UK\\
Work was done while an AI Resident at X \\Email: cb2015@cam.ac.uk}
\authorblockA{\authorrefmark{2}X, Mountain View, California, USA}
\authorblockA{\authorrefmark{3}Google Brain, Mountain View, California, USA}}

\maketitle


\begin{abstract}
The distributional perspective on reinforcement learning (RL) has given rise to a series of successful Q-learning algorithms, resulting in state-of-the-art performance in arcade game environments. 
However, it has not yet been analyzed how these findings from a discrete setting translate to complex practical applications characterized by noisy, high dimensional and continuous state-action spaces. 
In this work, we propose Quantile QT-Opt (Q2-Opt), a distributional variant of the recently introduced distributed Q-learning algorithm \cite{kalashnikov2018scalable} for continuous domains, and examine its behaviour in a series of simulated and real vision-based robotic grasping tasks. 
The absence of an actor in Q2-Opt allows us to directly draw a parallel to the previous discrete experiments in the literature without the additional complexities induced by an actor-critic architecture. 
We demonstrate that Q2-Opt achieves a superior vision-based object grasping success rate, while also being more sample efficient. 
The distributional formulation also allows us to experiment with various risk distortion metrics that give us an indication of how robots can concretely manage risk in practice using a Deep RL control policy. As an additional contribution, we perform batch RL experiments in our virtual environment and compare them with the latest findings from discrete settings. Surprisingly, we find that the previous batch RL findings from the literature obtained on arcade game environments do not generalise to our setup.
\end{abstract}

\IEEEpeerreviewmaketitle

\begin{figure*}[ht]
    \centering
    \includegraphics[width=0.75\textwidth]{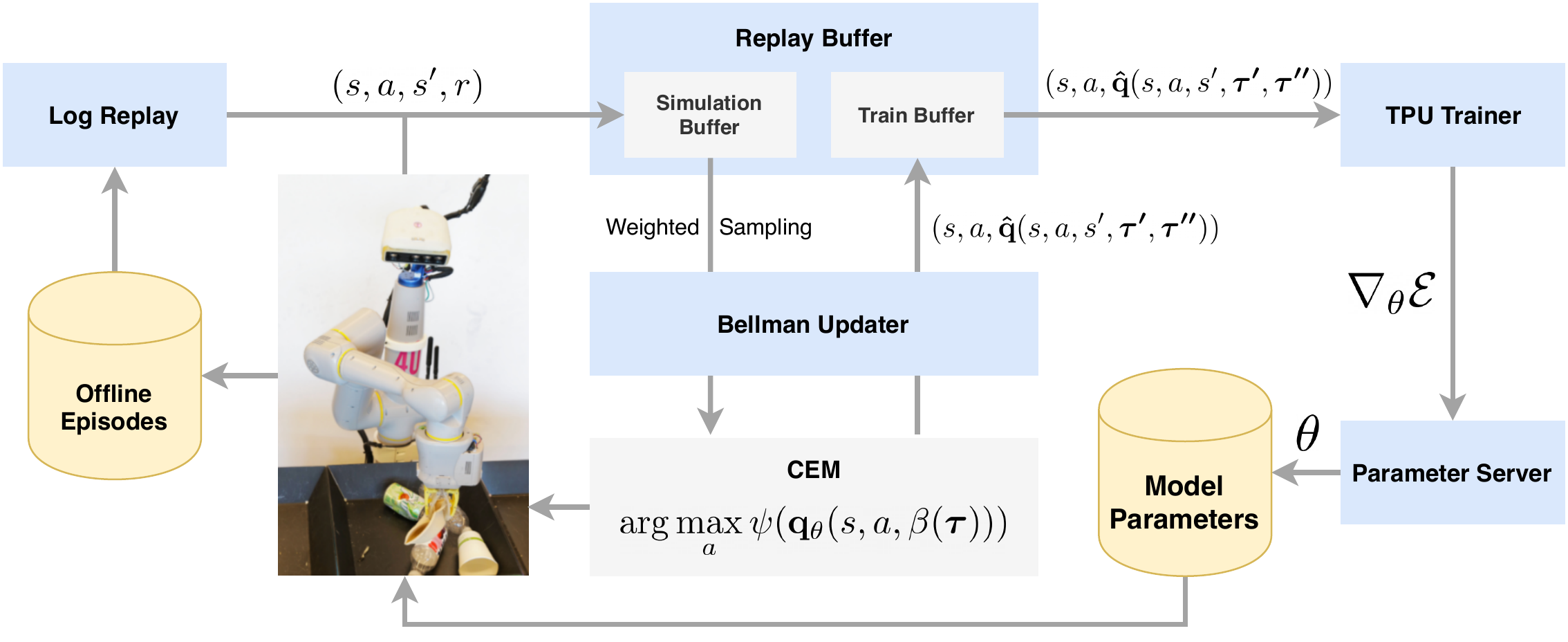}
    \caption{Distributed system architecture of Q2-Opt. The interactions between the robot and the environment are either stored in a database of episodes for later offline learning, or they are directly sent to the replay buffer when online learning is performed. The samples from the Simulation Buffer are pulled by the Bellman Updater, which appends the distributional targets. These labelled transitions are pushed to the train buffer and consumed by the TPU Training workers to compute the gradients. The parameter server uses the gradients to update the weights, which are asynchronously pulled by the agents.}
    \label{fig:q2t_opt_arch}
\end{figure*}

\section{Introduction}

The new distributional perspective on RL has produced a novel class of Deep Q-learning methods that learn a distribution over the state-action returns, instead of using the expectation given by the traditional value function. These methods, which obtained state-of-the-art results in the arcade game environments~\cite{bellemare2017distributional, dabney2018distributional, dabney2018implicit}, present several attractive properties. 

First, their ability to preserve the multi-modality of the action values naturally accounts for learning from a non-stationary policy, most often deployed in a highly stochastic environment. This ultimately results in a more stable training process and improved performance and sample efficiency. Second, they enable the use of risk-sensitive policies that no longer select actions based on the expected value, but take entire distributions into account. These policies can represent a continuum of risk management strategies ranging from risk-averse to risk-seeking by optimizing for a broader class of risk metrics. 

Despite the improvements distributional Q-learning algorithms demonstrated in the discrete arcade environments, it is yet to be examined how these findings translate to practical, real-world applications.
Intuitively, the advantageous properties of distributional Q-learning approaches should be particularly beneficial in a robotic setting. The value distributions can have a significant qualitative impact in robotic tasks, usually characterized by highly-stochastic and continuous state-action spaces. Additionally, performing safe control in the face of uncertainty is one of the biggest impediments to deploying robots in the real world, an impediment that RL methods have not yet tackled. In contrast, a distributional approach can allow robots to learn an RL policy that appropriately quantifies risks for the task of interest.

However, given the brittle nature of deep RL algorithms and their often counter-intuitive behaviour~\cite{Henderson2017DeepRL}, it is not entirely clear if these intuitions would hold in practice. Therefore, we believe that an empirical analysis of distributional Q-learning algorithms in real robotic applications would shed light on their benefits and scalability, and provide essential insight for the robot learning community. 

In this paper, we aim to address this need and perform a thorough analysis of distributional Q-learning algorithms in simulated and real vision-based robotic manipulation tasks. To this end, we propose a distributional enhancement of QT-Opt~\cite{kalashnikov2018scalable} subbed Quantile QT-Opt (Q2-Opt). 
The choice of QT-Opt, a recently introduced distributed Q-learning algorithm that operates on continuous action spaces, is dictated by its demonstrated applicability to large-scale vision-based robotic experiments. 
In addition, by being an actor-free generalization of Q-learning in continuous action spaces, QT-Opt enables a direct comparison to the previous results on the arcade environments without the additional complexities and compounding effects of an actor-critic-type architecture. 

In particular, we introduce two versions of Q2-Opt, based on Quantile Regression DQN (QR-DQN)~\cite{dabney2018distributional} and Implicit Quantile Networks (IQN)~\cite{dabney2018implicit}. The two methods are evaluated on a vision-based grasping task in simulation and the real world. We show that these distributional algorithms achieve state-of-the-art grasping success rate in both settings, while also being more sample efficient. Furthermore, we experiment with a multitude of risk metrics, ranging from risk-seeking to risk-averse, and show that risk-averse policies can bring significant performance improvements. We also report on the interesting qualitative changes that different risk metrics induce in the robots' grasping behaviour. As an additional contribution, we analyze our distributional methods in a batch RL scenario and compare our findings with an equivalent experiment from the arcade environments~\cite{Agarwal2019StrivingFS}.

\section{Related Work}

Deep learning has shown to be a useful tool for learning visuomotor policies that operate directly on raw images. Examples include various manipulation tasks, where related approaches use either supervised learning to predict the probability of a successful grasp~\cite{mahler2017dex, levine2016end, Pinto2015SupersizingSL} or learn a reinforcement-learning control policy \cite{levine2018learning, quillen2018deep, kalashnikov2018scalable}. 

Distributional Q-learning algorithms have been so far a separate line of research, mainly evaluated on game environments.
These algorithms replace the expected return of an action with a distribution over the returns and mainly vary by the way they parametrize this distribution. 
Bellemare et al. \cite{bellemare2017distributional} express it as a categorical distribution over a fixed set of equidistant points. Their algorithm, C51, minimizes the KL-divergence to the projected distributional Bellman target. A follow-up algorithm, QR-DQN \cite{dabney2018distributional}, approximates the distribution by learning the outputs of the quantile function at a fixed set of points, the quantile midpoints. This latter approach has been extended by IQN \cite{dabney2018implicit}, which reparametrized the critic network to take as input any probability $\tau$ and learn the quantile function itself. Besides this extension, their paper also analyses various risk-sensitive policies that the distributional formulation enables. In this work, we apply these advancements to a challenging vision-based real-world grasping task with a continuous action-space.

Closest to our work is D4PG~\cite{barth2018distributed}, a distributed and distributional version of DDPG~\cite{Lillicrap2015ContinuousCW} that achieves superior performance to the non-distributional version in a series of simulated continuous-control environments.
In contrast to this work, we analyze a different variant of Q-learning with continuous action spaces, which allows us to focus on actor-free settings that are similar to the previous distributional Q-learning algorithms. 
Besides, we demonstrate our results on real robots on a challenging vision-based grasping task.

\section{Background}

As previously stated, we build our method on top of QT-Opt~\cite{kalashnikov2018scalable}, a distributed Q-learning algorithm suitable for continuous action spaces. QT-Opt is one of the few scalable deep RL algorithms with demonstrated generalisation performance in a challenging real-world task. As the original paper shows, it achieves an impressive 96\% vision-based grasp success rate on unseen objects. Therefore, building on top of QT-Opt is a natural choice for evaluating value distributions on a challenging real-world task, beyond simulated environments. Additionally, by directly generalising Q-Learning to continuous domains, QT-Opt allows us to compare our results with the existing distributional literature on discrete environments.

In this paper, we consider a standard Markov Decision Process \cite{Puterman1994MarkovDP} formulation $(\mathcal{S}, \mathcal{A}, r, p, \gamma)$, where $s \in \mathcal{S}$ and $a \in \mathcal{A}$ denote the state and action spaces, $r(s, a)$ is a deterministic reward function, $p(\cdot|s, a)$ is the transition function and $\gamma \in (0, 1)$ is the discount factor. QT-Opt trains a parameterized state-action value function $Q_\theta(s, a)$ which is represented by a neural network with parameters $\theta$. The cross-entropy method (CEM)~\cite{Rubinstein2004TheCM} is used to iteratively optimize and select the best action for a given Q-function:
\begin{equation}
    \pi_\theta(s)=\argmax_{a}Q_\theta(s,a)
\end{equation}

In order to train the Q-function, a separate process called the ``Bellman Updater'' samples transition tuples $(s,a,r,s')$ containing the state $s$, action $a$, reward $r$, and next state $s'$ from a replay buffer and generates Bellman target values according to a clipped Double Q-learning rule~\cite{Hasselt2010DoubleQ,Sutton:1998:IRL:551283}:
\begin{equation}
    \hat{Q}(s, a, s') = r(s,a)+\gamma V(s') 
\end{equation}
where $V(s')=Q_{\bar{\theta}_1}(s',\pi_{\bar{\theta}_2}(s'))$, and $\bar{\theta_1}$ and $\bar{\theta_2}$ are the parameters of two delayed target networks. These target values are pushed to another replay buffer $\mathcal{D}$, and a separate training process optimizes the Q-function against a training objective:
\begin{equation}
\mathcal{E}(\theta)=\E_{(s,a,s')\sim\mathcal{D}}\left[D(Q_\theta(s,a),\hat{Q}(s, a, s'))\right]
\end{equation}
where $D$ is a divergence metric.

In particular, the cross-entropy loss is chosen for $D$, and the output of the network is passed through a sigmoid activation to ensure that the predicted Q-values are inside the unit interval.

\section{Quantile QT-Opt (Q2-Opt)}

In Q2-Opt (Figure \ref{fig:q2t_opt_arch}) the value function no longer predicts a scalar value, but rather a vector $\vq_\theta(s,a, \vtau)$ that predicts the quantile function output for a vector of input probabilities $\vtau$, with $\vtau_i \in [0, 1]$ and $  i=1,\ldots,N$. Thus the $i$-th element of $\vq_\theta(s,a, \vtau)$ approximates $F^{-1}_{s, a}(\vtau_i)$, where $F^{-1}_{s, a}$ is the inverse CDF of the random action-value associated with the state-action pair $(s, a)$. However, unlike QT-Opt where CEM optimizes directly over the Q-values, in Quantile QT-Opt, CEM maximizes a scoring function $\psi: \R^N \to \R$ that maps the vector $\vq$ to a score $\psi(\vq)$:
\begin{equation}
    \pi_\theta(s, \vtau)=\argmax_{a}\psi(\vq_\theta(s,a, \vtau))
\end{equation}
Similarly, the target values produced by the ``Bellman Updater'' are vectorized using a generalization of the clipped Double Q-learning rule from QT-Opt:
\begin{equation}
\begin{split}
    \hat{\vq}_{\bar{\theta}}(s, a, s', \vtau', \vtau'') &=r(s,a)\vone+\gamma\vv(s', \vtau', \vtau'') \\
    \vv(s', \vtau', \vtau'')&=\vq_{\bar{\theta}_1}(s',\pi_{\bar{\theta}_2}(s', \vtau''), \vtau')
\end{split}
\end{equation}
where $\vone$ is a vector of ones, and, as before, $\bar{\theta}_1$ and $\bar{\theta}_2$ are the parameters of two delayed target networks. Even though this update rule has not been considered so far in the distributional RL literature, we find it  effective in reducing the overestimation in the predictions. 

In the following sections, we present two versions of Q2-Opt based on two recently introduced distributional algorithms: QR-DQN and IQN. The main differences between them arise from the inputs $\vtau, \vtau'$, and $\vtau''$ that are used. To avoid overloading our notation, from now on we omit the parameter subscript in $\vq_\theta, \hat{\vq}_{\hat{\theta}}$ and replace it with an index into these vectors $\vq_i, \hat{\vq}_j$.

\subsection{Quantile Regression QT-Opt (Q2R-Opt)}

In Quantile Regression QT-Opt (Q2R-Opt), the vectors $\vtau, \vtau', \vtau''$ in $\vq$ and $\hat{\vq}$ are fixed. They all contain $N$ quantile midpoints of the value distribution. Concretely, $\vq_i(s,a,\vtau)$ is assigned the fixed quantile target $\vtau_i = \frac{\bar{\tau}_{i-1} + \bar{\tau}_{i}}{2}$ with $\bar{\tau}_i = \frac{i}{N}$. The scoring function $\psi(\cdot)$ takes the mean of this vector, reducing the $N$ quantile midpoints to the expected value of the distribution. Because $\vtau, \vtau', \vtau''$ are always fixed we consider them implicit and omit adding them as an argument to $\vq$ and $\hat{\vq}$ for Q2R-Opt.

The quantile heads are optimized by minimizing the Huber \cite{huber1964} quantile regression loss:
\begin{equation}
\begin{split}
     \rho_\tau^\kappa(\delta_{ij}) &= |\tau - \mathbb{I}\{\delta_{ij} < 0\}| \mathcal{L}_\kappa(\delta_{ij}) \\
     \mathcal{L}_\kappa(\delta_{ij}) &= 
     \begin{cases}
        \frac{1}{2} \delta_{ij}^2, & \text{if } |\delta_{ij}| \leq \kappa \\ 
        \kappa (|\delta_{ij}| - \frac{1}{2}\kappa),   & \text{otherwise} 
     \end{cases} 
\end{split}
\end{equation}
for all the pairwise TD-errors:
\begin{equation}
    \delta_{ij} = \hat{\vq}_j(s,a,s') - \vq_i(s, a)
\end{equation}
Thus, the network is trained to minimize the loss function:
\begin{equation}\label{eq:qh_loss}
\mathcal{E}(\theta)=\E_{(s,a,s')\sim\mathcal{D}}\left[\sum_{i=1}^N \E_j[\rho_{\vtau_i}^\kappa(\delta_{ij})]\right]
\end{equation}
We set $\kappa$, the threshold between the quadratic and linear regime of the loss, to  $0.002$ across all of our experiments. 

\subsection{Quantile Function QT-Opt (Q2F-Opt)}

In Q2F-Opt, the neural network itself approximates the quantile function of the value distribution, and therefore it can predict the inverse CDF for any $\vtau$. Since $\vtau, \vtau', \vtau''$ are no longer fixed, we explicitly include them in the arguments of $\vq$ and $\hat{\vq}$. Thus, the TD-errors $\delta_{ij}$ take the form:
\begin{equation}
    \delta_{ij} = \hat{\vq}_j(s, a, s', \vtau', \vtau'') - \vq_i(s, a, \vtau),
\end{equation}
where $\vtau_i \sim U[0, 1]$, $\vtau'_j \sim U[0, 1]$ and $\vtau''_j \sim U[0, 1]$ are sampled from independent uniform distributions. Using different input probability vectors also decreases the correlation between the networks. Note that now the length of the prediction and target vectors are determined by the lengths of $\vtau$ and $\vtau'$. The model is optimized using the same loss function as the one from Equation \ref{eq:qh_loss}.

\subsection{Risk-Sensitive Policies}

The additional information provided by a value distribution compared to the (scalar) expected return gives birth to a broader class of policies that go beyond optimizing for the expected value of the actions. Concretely, the expectation can be replaced with any risk metric, that is any function that maps the random return to a scalar quantifying the risk. In Q2-Opt, this role is played by the function $\psi$ that acts as a risk-metric. Thus the agent can handle the intrinsic uncertainty of the task in different ways depending on the specific form of $\psi$. It is important to specify that this uncertainty is generated by the environment dynamics $p(\cdot|s, a)$ and the (non-stationary) policy collecting the real robot rollouts, and that it is not a parametric uncertainty. 

We distinguish two methods to construct risk-sensitive policies for Q2R-Opt and Q2F-Opt, each specific to one of the methods. In Q2R-Opt, risk-averse and risk-seeking policies can be obtained by changing the function $\psi(\cdot)$ when selecting actions. Rather than computing the mean of the target quantiles, $\psi(\cdot)$ can be defined as a weighted average over the quantiles $\psi(\vq(s,a)) = \frac{1}{N} \sum w_i \vq_i(s,a)$. This sum produces a policy that is in between a worst-case and best-case action selector and, for most purposes, it would be preferable in practice over the two extremes. For instance, a robot that would consider only the worst-case scenario would most likely terminate immediately since this strategy, even though it is not useful, does not incur any penalty. Behaviours like this have been encountered in our evaluation of very conservative policies.

\begin{figure*}[ht]
    \centering
    \includegraphics[width=0.8\textwidth]{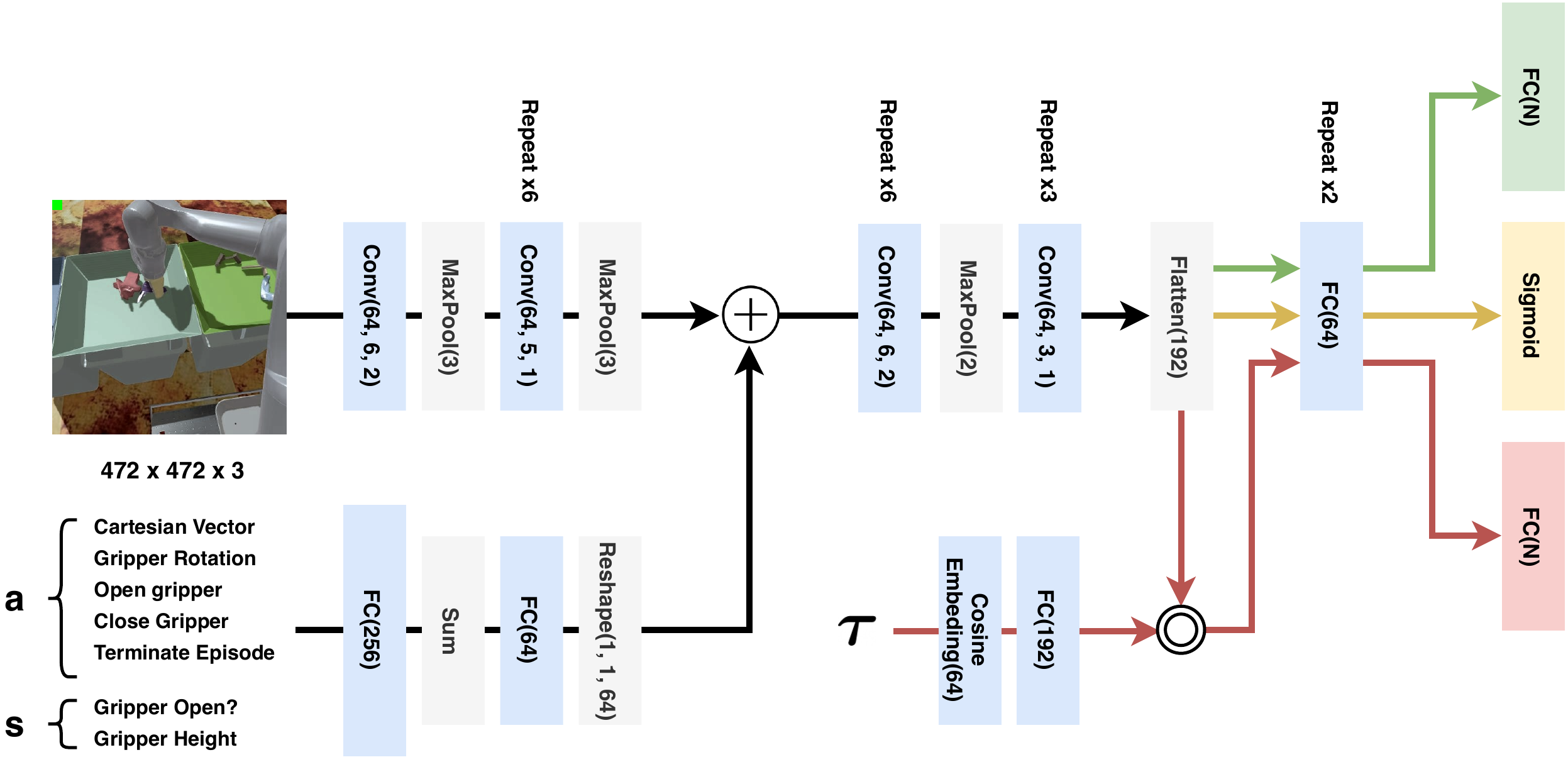}
    \caption{The neural network architectures for QT-Opt (yellow), Q2R-Opt (green) and Q2F-Opt (red). The common components of all the three models are represented by black arrows. The top-left image shows a view from the robot camera inside our simulation environment.}
    \label{fig:architectures}
\end{figure*}

In contrast, Q2F-Opt provides a more elegant way of learning risk-sensitive control policies by using risk distortion metrics \cite{wang_1996}. Recently, Majmidar and Pavone~\cite{Majumdar2017HowSA} have argued for the use of risk distortion metrics in robotics. They proposed a set of six axioms that any risk measure should meet to produce reasonable behaviour and showed that risk distortion metrics satisfy all of them. However, to the best of our knowledge, they have not been tried on real robotic applications. 

The key idea is to use a policy:
$$\pi_\theta(s, \beta(\vtau)) = \argmax_a \psi(\vq(s,a, \beta(\vtau))),$$ 
where $\beta: [0, 1] \to [0, 1]$ is an element-wise function that distorts the uniform distribution that $\vtau$ is effectively sampled from, and $\psi(\cdot)$ computes the mean of the vector as usual. Functions $\beta$ that are concave induce risk-averse policies, while convex function induce risk-seeking policies. 

\begingroup
\begin{table}[ht]
    \centering
    \renewcommand{\arraystretch}{1.7} 
    \begin{tabular}{l|c}
        \toprule
        Risk Metric & Formula \\ \midrule
        CPW & $\tau^\eta / (\tau^\eta + (1 - \tau)^\eta)^{\frac{1}{\eta}}$ \\
        Wang & $\Phi(\Phi^{-1}(\tau) + \eta)$ \\
        CVaR & $\eta \tau$ \\
        Norm & $\frac{1}{\eta} \sum_i^\eta \tau_i, \tau_i \sim U[0, 1]$ \\
        Pow & $\mathbb{I}_{\eta \geq 0} \tau ^ \frac{1}{1 + |\eta|} + \mathbb{I}_{\eta < 0} \Big[1 - (1 - \tau) ^ \frac{1}{1 + |\eta|}\Big]$ \\
        \bottomrule
    \end{tabular}
    \caption{The considered risk distortion metrics $\beta(\tau; \eta)$ with input $\tau$ and parametrised by $\eta$. $\Phi$ denotes the CDF of the standard normal distribution, and $\mathbb{I}$ is an indicator function.}
    \label{tab:risk_metrics}
\end{table}
\endgroup

In our experiments, we consider the same risk distortion metrics used by \citet{dabney2018implicit}: the cumulative probability weighting (\textbf{CPW})~\cite{Gonzlez1999OnTS}, the standard normal CDF-based metric proposed by \textbf{Wang}~\cite{Wang00aclass}, the conditional value at risk (\textbf{CVaR}) \cite{Rockafellar00optimizationof}, \textbf{Norm}~\cite{dabney2018implicit}, and a power law formula (\textbf{Pow})~\cite{dabney2018implicit}. Concretely, we use these metrics with a parameter choice similar to that of \citet{dabney2018implicit}. CPW(0.71) is known to be a good model for human behaviour \cite{Wu1996}; Wang$(-0.75)$, Pow$(-2)$, CVaR$(0.25)$ and CVaR$(0.4)$ are risk-averse. Norm($3$) decreases the weight of the distribution's tails by averaging 3 uniformly sampled $\vtau$. Ultimately, Wang$(0.75)$ produces risk-seeking behaviour. We include all these metrics in Table \ref{tab:risk_metrics}.

Due to the relationships between Q2F-Opt and the literature of risk distortion measures, we focus our risk-sensitivity experiments on the metrics mentioned above and leave the possibility of trying different functions $\psi(\cdot)$ in Q2R-Opt for future work. 



\subsection{Model Architecture}

To maintain our comparisons with QT-Opt, we use very similar architectures for Q2R-Opt and Q2F-Opt. For Q2R-Opt, we modify the output layer of the standard QT-Opt architecture to be a vector of size $N = 100$, rather than a scalar. For Q2F-Opt, we take a similar approach to ~\citet{dabney2018implicit}, and embed every $\vtau_k$ with $k \in \{1, \ldots, N = 32\}$ using a series of $n = 64$ cosine basis functions:
$$\phi_j(\vtau_k) := \text{ReLU}\left(\sum_{i=0}^{n-1} \cos(\pi i \vtau_k)w_{ij} + b_j\right)$$
We then perform the Hadamard product between this embedding and the convolutional features. Another difference in Q2F-Opt is that we replace batch normalization~\cite{Ioffe:2015:BNA:3045118.3045167} in the final fully-connected layers with layer normalization~\cite{Ba2016LayerN}. We notice that this better keeps the sampled values in the range allowed by our MDP formulation.  The three architectures are all included in a single diagram in Figure \ref{fig:architectures}. 

\section{Results}

In this section, we present our results on simulated and real environments. In simulation, we perform both online and offline experiments, while for the real world, the training is exclusively offline. We begin by describing our evaluation method. 

\subsection{Experimental Setup}

We consider the problem of vision-based robotic grasping for our evaluations. In our grasping setup, the robot arm is placed at a fixed distance from a bin containing a variety of objects and tasked with grasping any object. The MDP specifying our robotic manipulation task provides a simple binary reward to the agent at the end of the episode: $0$ for a failed grasp, and $1$ for a successful grasp. To encourage the robot to grasp objects as fast as possible, we use a time step penalty of $-0.01$ and a discount factor $\gamma=0.9$. The state is represented by a $472\times472$ RGB image; the actions are a mixture of continuous 4-DOF tool displacements in $x$, $y$, $z$ with azimuthal rotation $\phi$, and discrete actions to open and close the gripper, as well as to terminate the episode. 

In simulation, we train the agent to grasp from a bin containing 8 to 12 randomly generated procedural objects (Figure \ref{fig:architectures}). For the first $5,000$ global training steps, we use a procedural exploration policy. The scripted policy is lowering the end effector at a random position at the level of the bin and attempts to grasp. After $5,000$ steps, we switch to an $\epsilon$-greedy policy with $\epsilon = 0.2$. We train the network from scratch (no pretraining) using Adam \citep{kingma2014adam} with a learning rate $10^{-4}$ and batch size $4096$ ($256$ per chip on a $4\times4$ TPU). Additionally, we use two iterations of CEM with 64 samples for each. 

In the real world, we train our model offline from a $72$ TiB dataset of real-world experiences collected over five months, containing $559,642$ episodes of up to $20$ time steps each. Out of these, $39\%$ were generated by noise-free trained QT-Opt policies, $22\%$ by an $\eps$-greedy strategy using trained QT-Opt policies and $39\%$ by an $\eps$-greedy strategy based on a scripted policy.  For evaluation, we attempt 6 consecutive grasps from a bin containing 6 objects without replacement, repeated across 5 rounds. Figure \ref{fig:q2t_opt_arch} includes our workspace setup. We perform this experiment in parallel on 7 robots, resulting in a total of $210$ grasp attempts. All the robots use a similar object setup consisting of two plastic bottles, one metal can, one paper bowl, one paper cup, and one paper cup sleeve. In the results section, we report the success rate over the $210$ attempts. Videos for the real-world experiments can be found on the website\footnote{\url{https://q2-opt.github.io/}} accompanying the paper. 

Our evaluation methodology is different from that of \citet{kalashnikov2018scalable}. The original QT-Opt paper reports an average success rate of 76\% for grasping 28 objects over 30 attempts without replacement, trained on a mixture of off-policy and on-policy data. While not directly comparable, we reproduce QT-Opt on a completely different robot with different objects and report an average success rate of 70\% for grasping 6 objects over 6 attempts without replacement, trained on off-policy data only.

\subsection{Simulation Experiments}

\begin{figure}[ht]
    \centering
    \includegraphics[width=0.9\columnwidth]{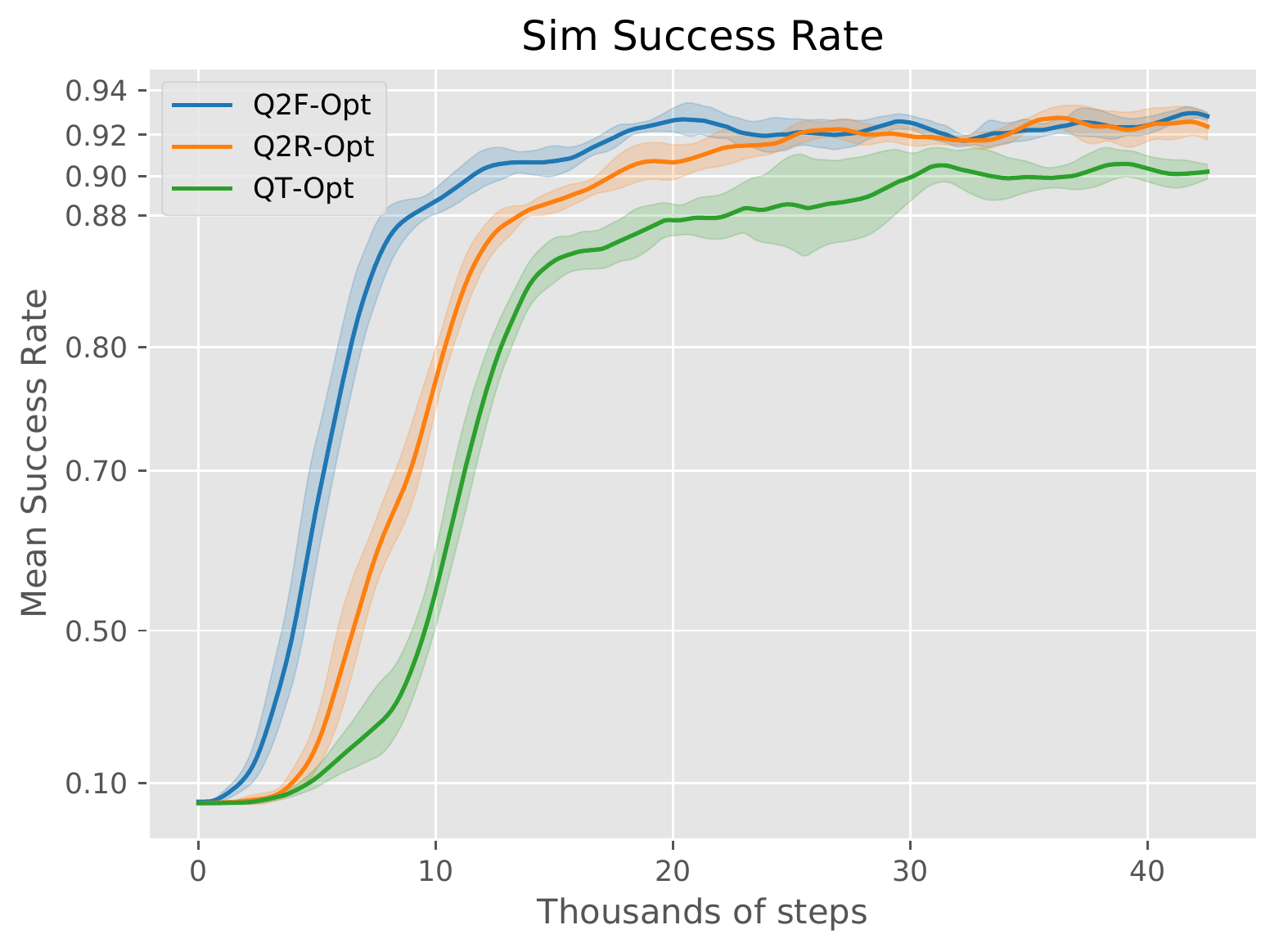}
    \caption{Sim success rate as a function of the global step. The distributional methods achieve higher grasp success rates in a lower number of global steps.}
    \label{fig:sim_success}
\end{figure}

\begin{figure}[ht]
    \centering
    \includegraphics[width=0.9\columnwidth]{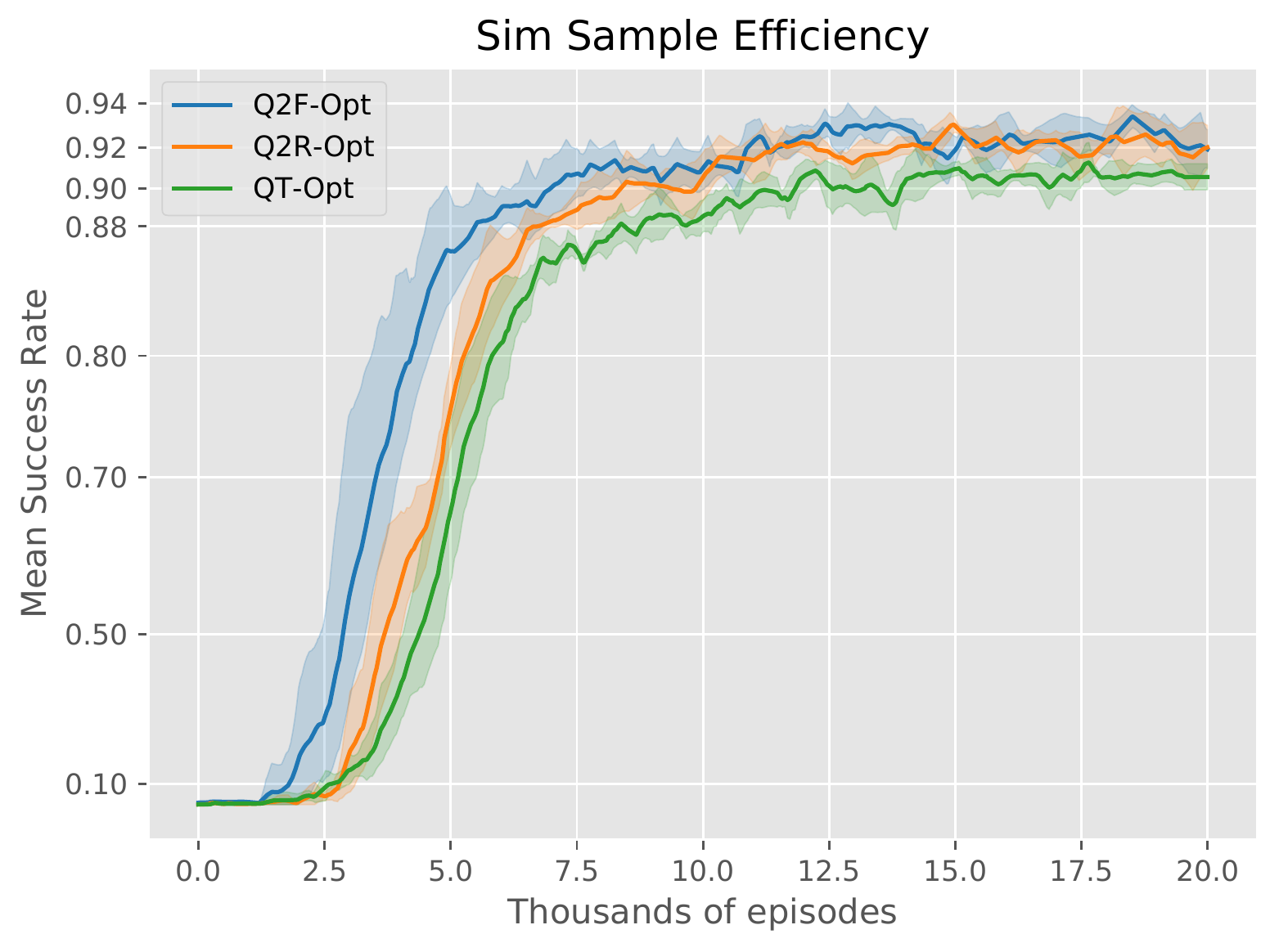}
    \caption{Sim success rate as a function of the number of generated environment episodes. The distributional methods are significantly more sample efficient than QT-Opt.}
    \label{fig:sim_sample_success}
\end{figure}

We begin by evaluating Q2-Opt against QT-Opt in simulation. Figure~\ref{fig:sim_success} shows the mean success rate as a function of the global training step together with the standard deviation across five runs for QT-Opt, Q2R-Opt and Q2F-Opt. Because Q2-Opt and QT-Opt are distributed systems, the global training step does not directly match the number of environment episodes used by the models during training. Therefore, to understand the sample efficiency of the algorithm, we also include in Figure~\ref{fig:sim_sample_success} the success rate as a function of the total number of environment episodes added to the buffer. 

The distributional methods achieve higher success rates while also being more sample efficient than QT-Opt. While Q2F-Opt performs best, Q2R-Opt exhibits an intermediary performance and, despite being less sample efficient than Q2F-Opt, it still learns significantly faster than our baseline. 


\begin{figure}[ht]
    \centering
    \includegraphics[width=0.9\columnwidth]{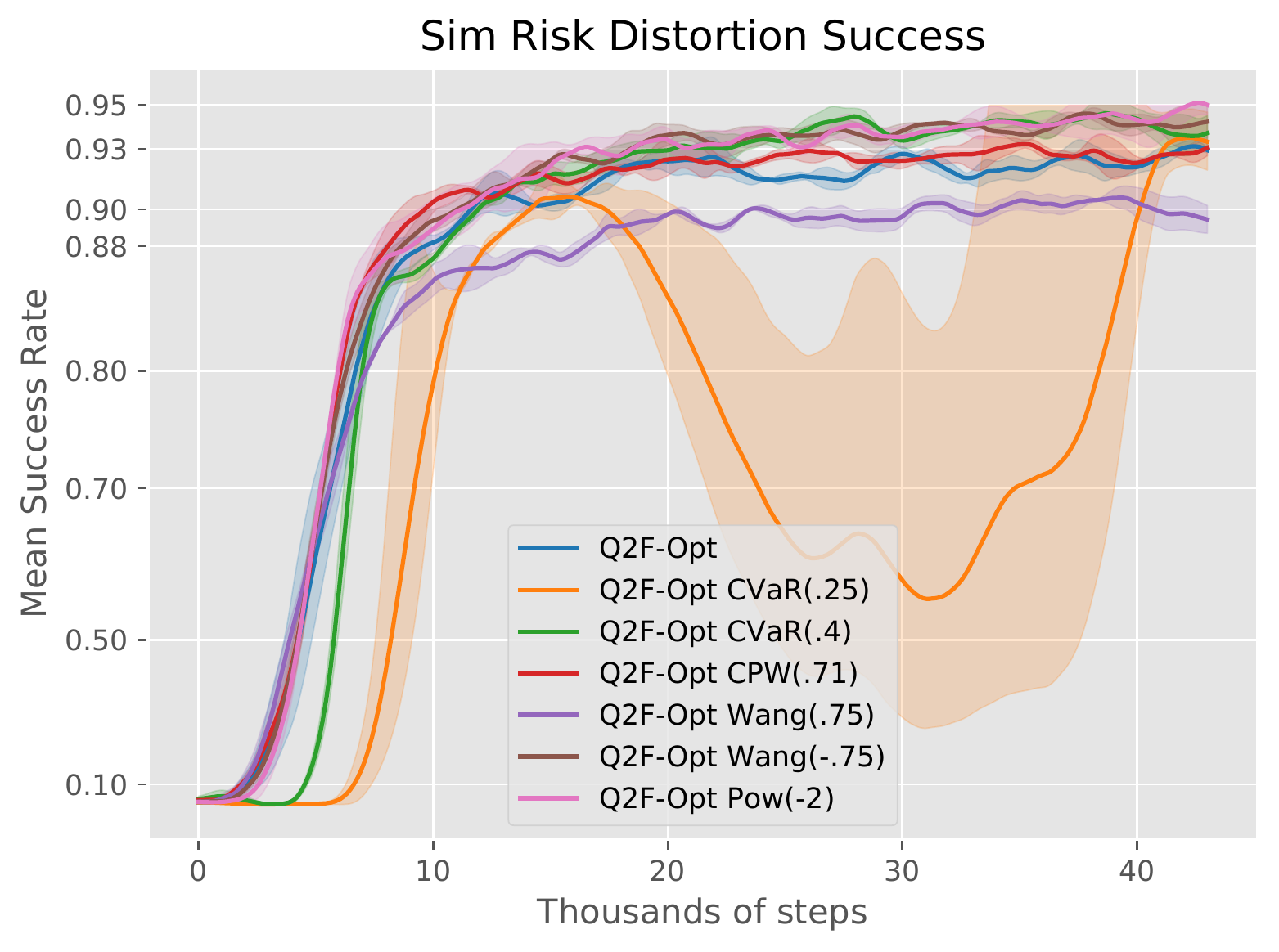}
    \caption{Average success rate over five runs for different risk-sensitive policies in sim. Most risk-averse policies perform better, but extremely conservative ones like CVaR$(0.25)$ can become unstable. The risk-seeking policy Wang$(0.75)$ performs worst.}
    \label{fig:sim_risk_success}
\end{figure}

We extend these simulation experiments with a series of risk distortion measures equipped with different parameters. Figure~\ref{fig:sim_risk_success} shows the success rate for various measures used in Q2F-Opt. We notice that risk-averse policies (Wang$(-0.75)$, Pow$(-2)$, CVaR) are generally more stable in the late stages of training and achieve a higher success rate. Pow$(-2)$ remarkably achieves 95\% grasp success rate. However, being too conservative can also be problematic. Particularly, the CVaR$(0.25)$ policy becomes more vulnerable to the locally optimal behaviour of stopping immediately (which does not induce any reward penalty). This makes its performance fluctuate throughout training, even though it ultimately obtains a good final success rate. Table \ref{tab:sim_success} gives the complete final success rate statistics. 

\begin{table}[ht]
    \centering
    \begin{tabular}{l|ccc}
        \toprule
        Model  & Success & Std & Median  \\ \midrule
        QT-Opt  & 0.903 & 0.005  & 0.903 \\
        Q2R-Opt & 0.923 & 0.006 & 0.924 \\
        Q2F-Opt & \textbf{0.928} & \textbf{0.001} & \textbf{0.928} \\ \midrule
        Q2F-Opt Wang(0.75)  & 0.898 & 0.012 & 0.893 \\
        Q2F-Opt CPW(0.71)  & 0.928 & 0.003 & 0.925 \\
        Q2F-Opt CVAR(0.25)   & 0.933 & 0.013 & 0.941 \\
        Q2F-Opt CVAR(0.4)  & 0.938 & 0.008 & 0.938 \\
        Q2F-Opt Wang(-0.75)  & 0.942 & 0.007 & 0.944 \\
        Q2F-Opt Pow(-2.0)  & \textbf{0.950} & \textbf{0.004} & \textbf{0.952} \\
        \bottomrule
    \end{tabular}
    \caption{Final sim success rate statistics. Distributional risk-averse policies have the best performance.}
    \label{tab:sim_success}
\end{table}

\subsection{Real-World Experiments}

The chaotic physical interactions specific to real-world environments and the diversity of policies used to gather the experiences make the real environment an ideal setting for distributional RL. Furthermore, this experiment is also of practical importance for robotics since any increase in grasp success rate from offline data reduces the amount of costly online training that has to be performed to obtain a good policy. 

\begin{table}[ht]
    \centering
    \begin{tabular}{l|c}
        \toprule
        Model & Grasp Success Rate \\ \midrule
        QT-Opt & 70.00\% \\
        Q2R-Opt & 79.50\% \\
        Q2F-Opt & \textbf{82.00\%} \\ \midrule
        Q2F-Opt CPW(0.71) & 75.71\% \\
        Q2F-Opt Wang(0.75) & 78.10\% \\
        Q2F-Opt Norm(3) & 80.47\% \\
        Q2F-Opt Pow(-2) & 83.81\% \\
        Q2F-Opt CVaR(0.4) & 85.23\% \\ 
        Q2F-Opt Wang(-0.75) & \textbf{87.60\%} \\
        \bottomrule
    \end{tabular}
    \caption{Real world grasp success rate out of $210$ total grasps. Our methods significantly outperform QT-Opt, while risk-averse polices are better by a significant margin.}
    \label{tab:real_results}
\end{table}

We report in Table \ref{tab:real_results} the grasp success rate statistics for all the considered models. We find that the best risk-averse version of Q2-Opt achieves an impressive 17.6\% higher success rate than QT-Opt. While the real evaluation closely matches the model hierarchy observed in sim, the success rate differences between the models are much more significant. 

\begin{figure}[!h]
    \centering
    \includegraphics[width=0.9\columnwidth]{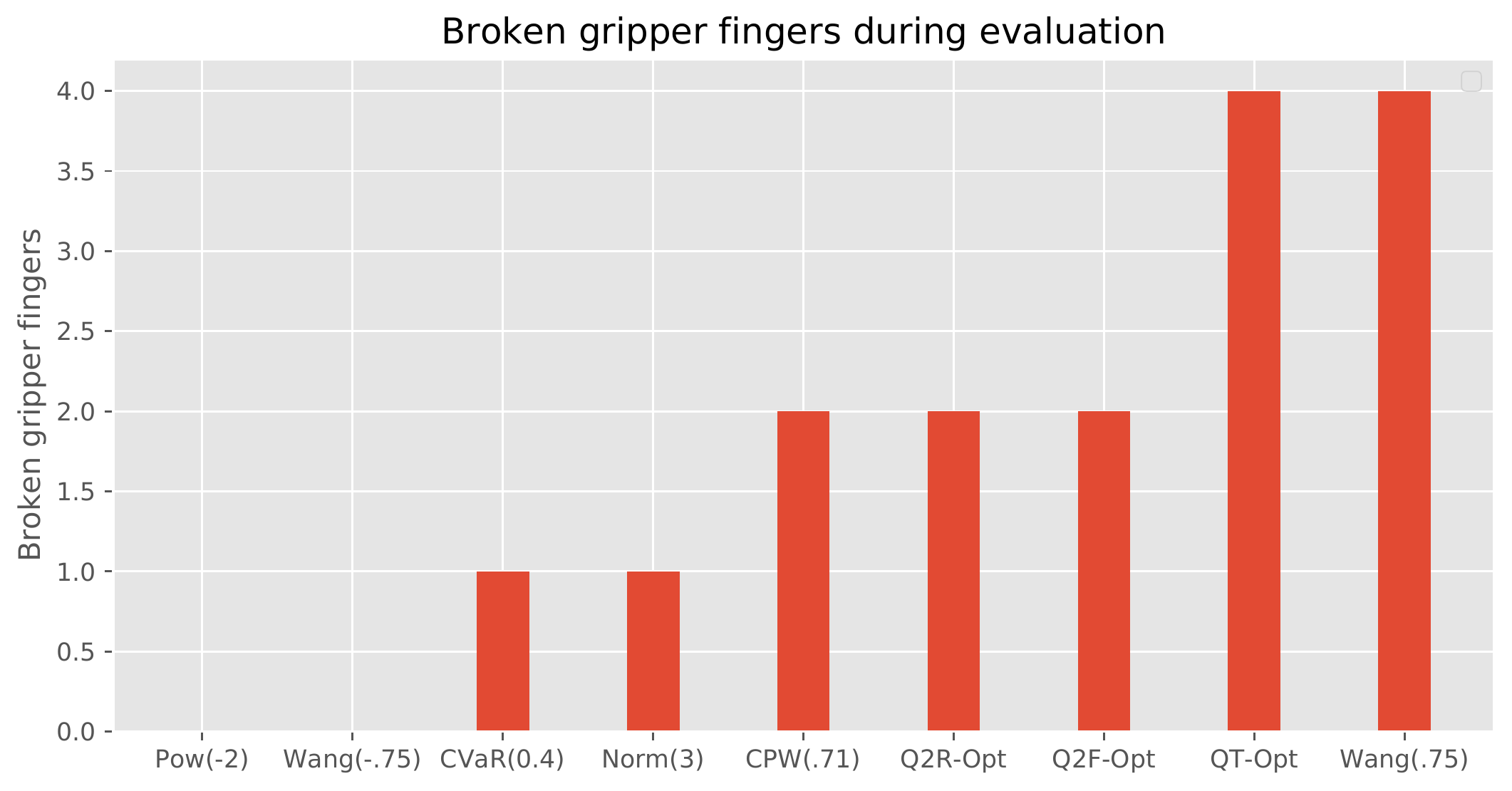}
    \caption{The number of detached gripper fingers during evaluation. Among the risk-averse policies (first three), only one gripper broke. The agents controlled by risk-neutral policies (Q2R-Opt, Q2F-Opt, QT-Opt) lost eight gripper fingers in total, with half of those belonging to QT-Opt. The last policy, which is risk-seeking lost four, similar to QT-Opt. The other policies (Norm, CPW) behaved similarly to risk-neutral policies.}
    \label{fig:broken_fingers}
\end{figure}

\begin{figure*}[!ht]
    \centering
        \begin{subfigure}[t]{0.30\textwidth}
         \centering
         \includegraphics[width=\columnwidth]{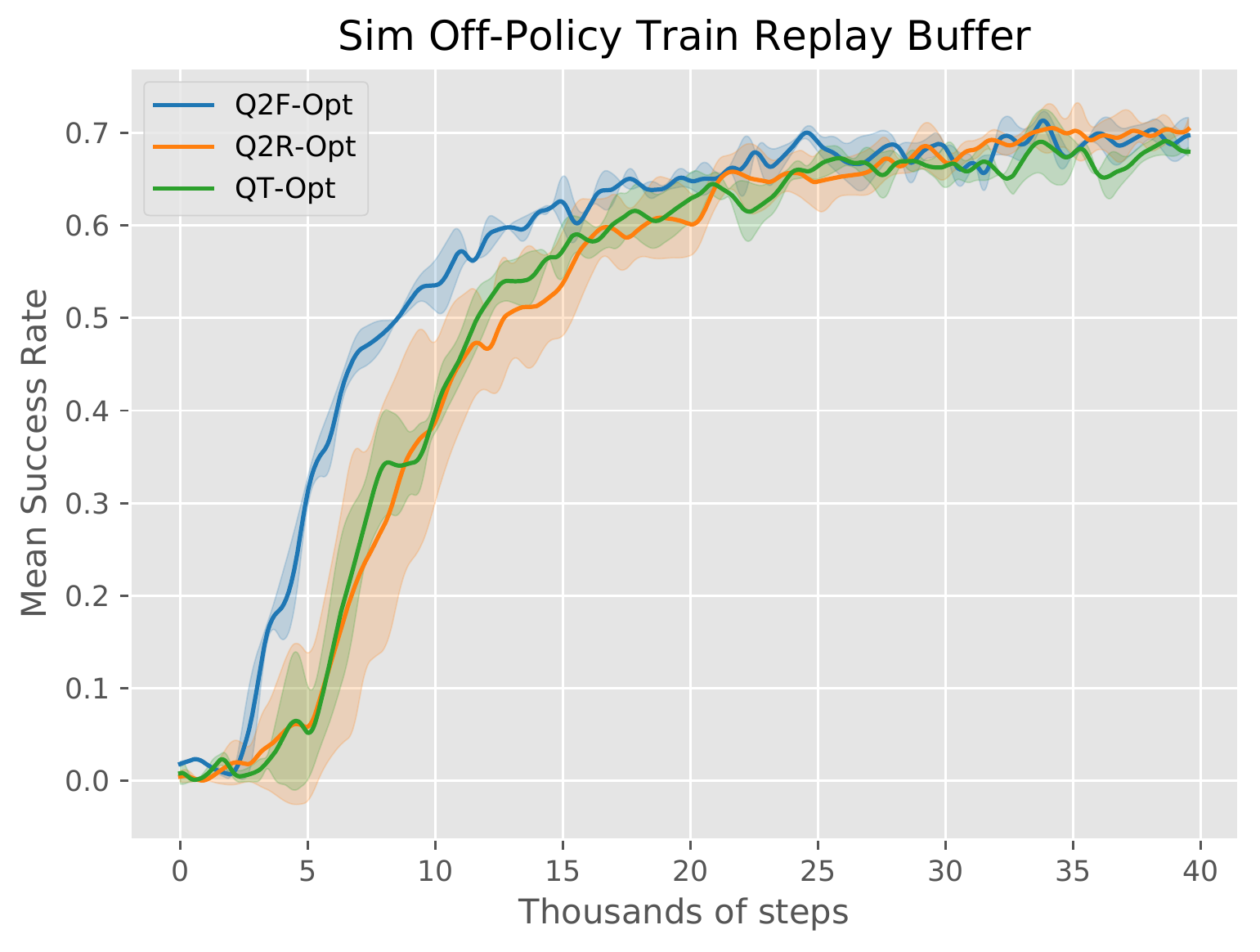}
         \caption{Sim off-policy success rate for data collected during a full training run. None of the methods can achieve the final performance of the policy trained online (90\%).}
         \label{fig:sim_offp_train_success}
    \end{subfigure}
    ~
    \begin{subfigure}[t]{0.30\textwidth}
         \centering
         \includegraphics[width=\columnwidth]{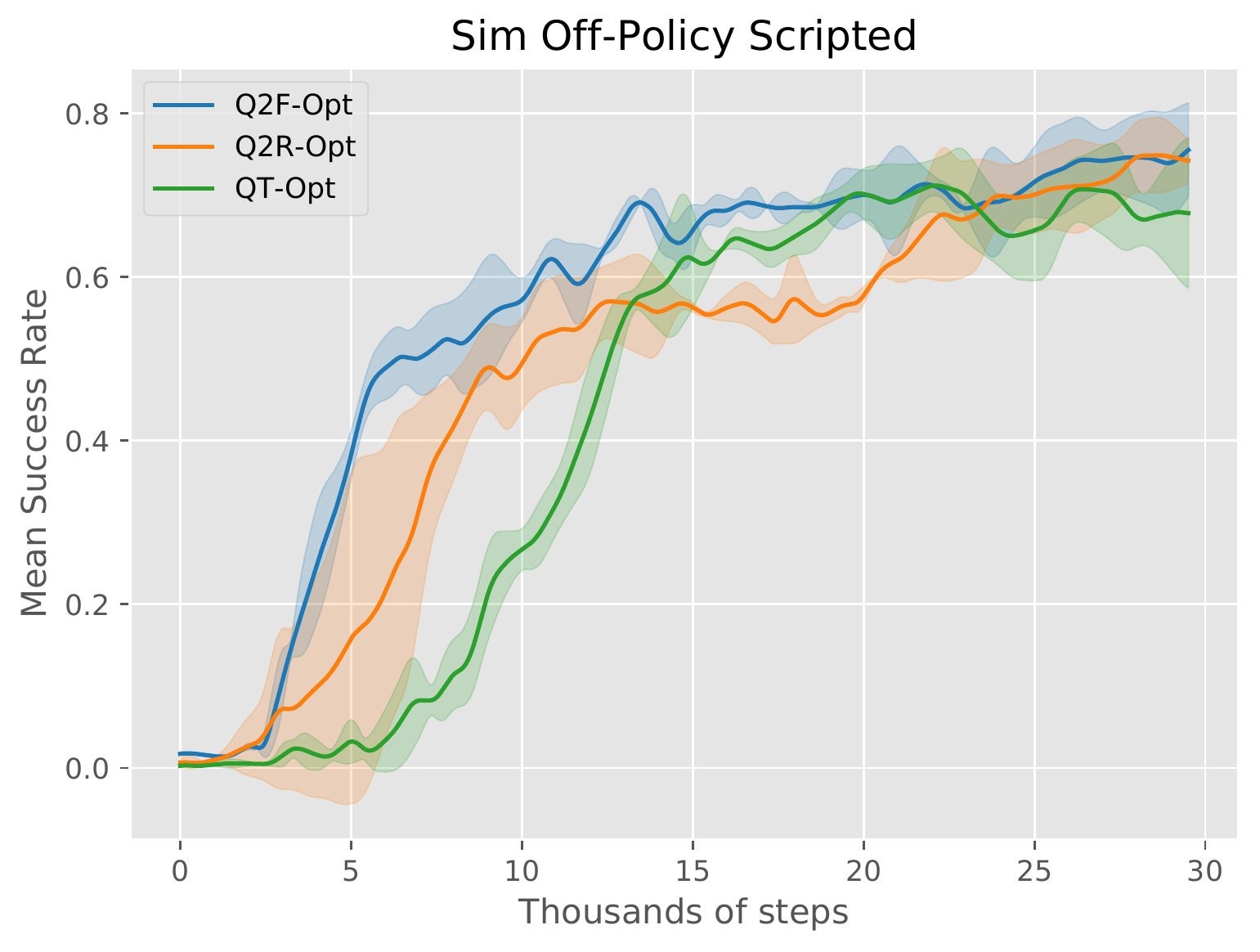}
         \caption{Sim success rate from an offline dataset produced by a scripted exploration policy. The models achieve a higher success rate than on the replay buffer dataset.}
         \label{fig:sim_scripted}
    \end{subfigure}
    ~ 
    \begin{subfigure}[t]{0.31\textwidth}
        \centering
        \includegraphics[width=\columnwidth]{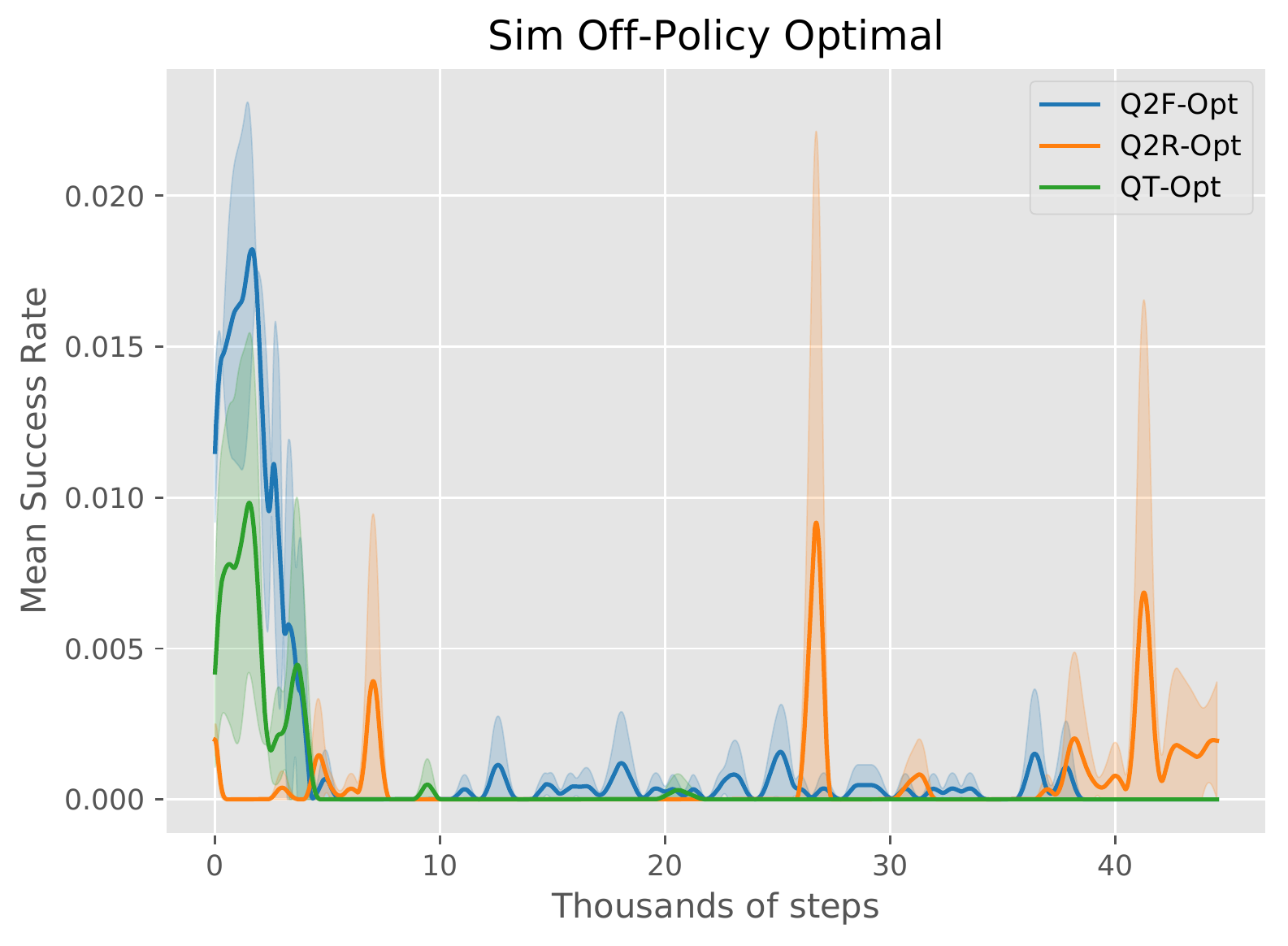}
        \caption{Sim success rate from an offline dataset produced by a policy close to optimality. None of the models is able to learn from a dataset of successful grasps.}
        \label{fig:sim_optimal}
    \end{subfigure}
    \caption{Batch RL experiments in simulation. Distributional methods show little to no improvement in our simulated environment.}
\end{figure*}

Besides the improvement in performance, we notice that the distortion measures of Q2F-Opt have a significant qualitative impact, even though the training is performed from the same data. Risk-averse policies tend to readjust the gripper in positions that are more favourable or move objects around to make grasping easier. CVaR(0.4), the most conservative metric we tested in the real world, presented a particularly interesting behaviour of intentionally dropping poorly grasped objects to attempt a better re-grasp. The CVaR policy mainly used this technique when attempting to grasp objects from the corners of the bin to move them in a central position.

However, a downside of risk-averse policies that we noticed is that, for the difficult-to-grasp paper cup sleeves, the agent often kept searching for an ideal position without actually attempting to grasp. We believe this is an interesting example of the trade-offs between being conservative and risk-seeking. The only tested risk-seeking policy, using Wang(0.75), made many high-force contacts with the bin and objects, which often resulted in broken gripper fingers and objects being thrown out of the bin. We also showcase some of these qualitative differences between the considered policies in the videos accompanying our paper. 

These qualitative differences in behaviour that value distributions cause can provide a way to achieve safe robot control. An interesting metric we considered to quantify these behaviours is the number of broken gripper fingers throughout the entire evaluation process presented above. Occasionally, the gripper fingers break in high-force contacts with the objects or the bin. Figure \ref{fig:broken_fingers} plots these numbers for each policy. Even though we do not have a statistically significant number of samples, we believe this figure is a good indicator that risk-averse policies implicitly achieve safer control. 

\subsection{Batch RL and Exploitation}

Recently, Agarwal et al.~\cite{Agarwal2019StrivingFS} have argued that most of the advantages of distributional algorithms come from better exploitation. Their results demonstrated that QR-DQN could achieve in offline training a performance superior to online C51. Since environment interactions are particularly costly in robotics, we aim to reproduce these results in a robotic setting. Therefore, we perform an equivalent experiment in simulation and train the considered models on all the transitions collected during training by a QT-Opt agent with a final success rate of 90\% (Figure \ref{fig:sim_offp_train_success}). 

We note that despite the minor success rate improvements brought by Q2R-Opt and Q2F-Opt, the two models are not even capable of achieving the final success rate of the policy trained from the same data. We hypothesize this is due to the out-of-distribution action problem~\cite{Kumar2019StabilizingOQ}, which becomes more prevalent in continuous action spaces. 

We investigated this further on two other datasets: one collected by a scripted stochastic exploration policy with $46\%$ success rate and another produced by an almost optimal policy with $89\%$ grasp success rate. Figures  \ref{fig:sim_scripted} and \ref{fig:sim_optimal} plot the results for these two datasets. Surprisingly, the models achieve a higher success rate on the scripted exploration dataset than on the dataset collected during training. On the dataset generated by the almost optimal policy, none of the methods manages to obtain a reasonable success rate. These results, taken together with our real-world experiments, suggest that offline datasets must contain a diverse set of experiences to learn effectively in a batch RL setting.

\section{Conclusion}

In this work, we have examined the impact that value distributions have on practical robotic tasks. Our proposed methods, collectively called Q2-Opt, achieved state-of-the-art success rates on simulated and real vision-based robotic grasping tasks, while also being significantly more sample efficient than the non-distributional equivalent, QT-Opt. Additionally, we have shown how safe reinforcement learning control can be achieved through risk-sensitive policies and reported the rich set of behaviours these policies produce in practice despite being trained from the same data. As a final contribution, we evaluated the proposed distributional methods in a batch RL setting similar to that of Agarwal et al.~\cite{Agarwal2019StrivingFS} and showed that, unfortunately, their findings do not translate to the continuous grasping environment presented in this work.  

\section*{Acknowledgments}

We would like to give special thanks to Ivonne Fajardo and Noah Brown for overseeing the robot operations. We would also like to extend our gratitude to Julian Ibarz for helpful comments. 


\newpage
\bibliographystyle{plainnat}
\bibliography{references}

\end{document}